# Robust Anomaly-Based Ship Proposals Detection from Pan-sharpened High-Resolution Satellite Image

Viet Hung Luu, Nguyen Hoang Hoa Luong, Quang Hung Bui, and Thi Nhat Thanh Nguyen

*Abstract*—Pre-screening of ship proposals is now employed by top ship detectors to avoid exhaustive search across image. In very high resolution (VHR) optical image, ships appeared as a cluster of abnormal bright pixels in open sea clutter (noise-like background). Anomaly-based detector utilizing Panchromatic (PAN) data has been widely used in many researches to detect ships, however, still facing two main drawbacks: 1) detection rate tend to be low particularly when a ship is low contrast and 2) these models require a high manual configuration to select a threshold value best separate ships from sea surface background. This paper aims at further investigation of anomaly-based model to solve those issues. First, pan-sharpened Multi Spectral (MS) data is incorporated together with PAN to enhance ship discrimination. Second, we propose an improved anomaly-based model combining both global intensity anomaly and local texture anomaly map. Regarding noise appeared due to the present of sea clutter and because of pan-sharpen process, texture abnormality suppression term based on quantization theory is introduced. Experimental results on VNREDSat-1 VHR optical satellite images suggest that the pan-sharpened near-infrared (P-NIR) band can improve discrimination of ships from surrounding waters. Compared to state-of-the-art anomaly-based detectors, our proposed anomaly-based model on the combination of PAN and P-NIR data cannot only achieved highest ship detection's recall rate (91.14% and 45.9% on high-contrast and low-contrast dataset respectively) but also robust to different automatic threshold selection techniques.

## I. INTRODUCTION

MARINE ship monitoring in coastal region is an increasingly important task for the safety and security of maritime traffic. The International Maritime Organization's International Convention for the Safety of Life at Sea requires Automatic Identification System (AIS) to be fitted aboard international voyaging ships with 300 or more gross tonnage. AIS provides ship unique identification such as name, details, location, speed, and heading which are transmitted frequently to the ground center. However, the AIS might be purposely switched off, defected or simply not equipped for small ships [1]. To prevent illegal activities on the waters, e.g. illegal fishery, pollution, immigration, it needs another collaborating monitor system which based on remote sensing data to locate ships without functioning AIS.

Synthetic Aperture Radar (SAR) and high resolution optical images are widely used in operationally. SAR images are less affected by weather conditions and can be utilized to estimate velocity of ship target. However, they are usually with high level speckles and difficult for human interpretation [2]. Ship detection on optical satellite images can extend the SAR based systems. The main advantage of optical satellite images is that they can have very high spatial resolution, thus enabling the detection of small ships and enhancing further ship type recognition.

*A. Related Work*

Existing works on ship candidates' selection can be divided into three main groups. The first group performs pixel wise labeling to address the foreground pixels and then group them into regions by incorporating region growing approach. These methods focus on the difference in gray values between foreground object including ships and other inferences such as clouds, wake … and background sea surface. A threshold segment method is applied to produce the binary image and then post-processed using morphological operators to remove noises and connect components. This approach has a major problem. Since the lack of prior analysis on sea surface model, parameters and threshold values of these methods are usually empirical chosen, which lacks the robustness. They may either over segment the ship into small parts or make the ship candidate merge to nearby land or cloud regions [3]. Corbane et al. 2008 [4] was the first to develop a method for the detection of ships using the contrast between ships and background of panchromatic (PAN) image. In [2], the idea of incorporating sea surface analysis to ship detection using PAN image was first declared. They defined two novel features to describe the intensity distribution of majority and effective pixels. The two features cannot only quickly block out no-candidate regions, but also measure the Intensity Discrimination Degree of the sea surface to assign weights for ship candidate selection function automatically. In [5], the authors re-arrange the spatially adjacent pixels into a vector,

This work was supported by the Space Technology Program of Vietnam under Grant VT-UD/06/16-20.

Viet Hung Luu, Quang Hung Bui and Thi Nhat Thanh Nguyen are with Center of Multidisciplinary Integrated Technologies for Field Monitoring, University of Engineering and Technology – Vietnam National University, Vietnam (email: {hunglv, thanhntn, hungbq}@fimo.edu.vn). Nguyen Hoang Hoa Luong is with University of Engineering and Technology – Vietnam National University, Vietnam (email: hoalnh79@gmail.com).

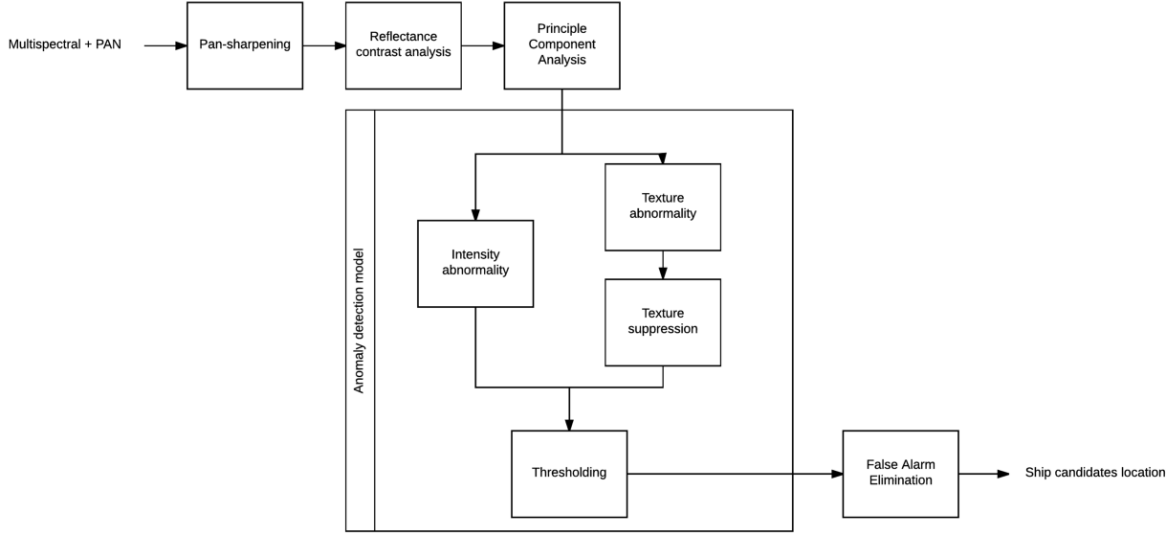

Fig. 1. Diagram of proposed ship detection approach, which can be separated into two main parts: spectral data analysis and ship proposals extraction.

transforming the panchromatic image into a "fake" hyper-spectral form. The hyper-spectral anomaly detection named RXD [6], [7] was applied to extract ship candidates efficiently, particularly for the sea scenes which contain large areas of smooth background.

The methods in second group have incorporated bounding box labeling. [1], [8], [9] detected ships based on sliding windows in varying sizes. However, only labeling bounding boxes is not accurate enough for ship localization; thus, it is unsuited for ship classification [10]. In [11], [12], the authors detected ships by shape analysis, including ship head detection after water and land segmentation and removed false alarms by labeling rotated bounding box candidates. These methods depend heavily on detecting of V-shape ship heads which is not applicable for small-size ship detection in low resolution images (2.5 meters or lower). In [10] the authors proposed ship rotated bounding box which is the improvement of the second group. Ship rotated bounding box space using modified version of BING object-ness score [13] is defined which reduce the search space significant. However, this method has low Average Recall in compare to pixel-wise labeling methods.

*B. Contributions*

In this paper, we aim to further investigate the problems arising from anomaly-based ship detector using PAN image. In the pre-detection stage, we first pan-sharpen Multi-Spectral (MS) bands to have the same spatial resolution as PAN image. Bands, which best discriminate ships from surrounding background, are integrated into a single band image using principle component analysis (PCA). Then, an anomaly detector based on texture and intensity abnormality is applied to extract ship candidates. Our contributions are threefold.

First, to the best of our knowledge, the visible and near infrared bands of Very High Resolution (VHR) images for ship detection has so far not been investigated together due to its inferior spatial resolution. By applying pan-sharpening method, we can analyze the performance of pan-sharpened MS bands and PAN for detecting ships.

Second, we introduce texture suppression term which help project the abnormal pixels as ship while suppressing ones appeared as white noise in the anomaly detection model. Combined with the integrated PCA data, the proposed algorithm cannot only extract ship candidates efficiently but also robustness to different automatic thresholding technique.

Third, a comprehensive evaluation with state-of-the-art anomaly-based methods is provided. In this work, we aim to revisit existing works and compare most publicly available methods in a unified framework. It allows us to better understand the benefits and limitations of our proposed anomaly-based detector versus others.

The rest of this paper is organized as follows. In Section II, the complete approach for ship proposals extraction is given. In Section III, the execution of proposed approach is illustrated. Quantitative comparison and evaluation of PAN and MS bands and our overall approach for extracting ship proposals are provided. Finally, the conclusion is discussed in Section IV.

II. SHIP PROPOSALS EXTRACTION

The methodological framework is shown in Fig. 1 which consists of two main stages. In the first stage, the MS images with lower spatial resolution are first pan-sharpened into new images with more spatial details while remaining spectral information using UNB method [14]. Then the reflectance contrast [15] between ships and the adjacent region (water and interferences) is introduced to assess the capability of ship identification of each pan-sharpened MS images and PAN image. Bands which better discriminate ships from surrounding waters are fed into principle component analysis (PCA) module. The first PCA component which explains most of the data variation is then used as the input of ship proposals extraction stage. Anomaly detection model based on texture and intensity anomaly derived from edge operator and frequency of grey-level value respectively is employed. Ship proposals are extracted by an automatic threshold on anomaly image. To make the automatic threshold algorithm more robust to image with noisy sea background, we introduce texture suppression term which suppresses the small texture values. Finally, simple linear false alarm elimination is employed to quickly and preliminarily remove obvious false alarm. In this paper, our study aims at detecting ships in open water. The land area can be masked out using prior geographic information.

*A. Pan-sharpening MS and PAN data*

Due to cost and complexity issues, recently launched VHR satellites often provide us a PAN image with finer spatial resolution than MS images. However, MS images have higher spectral resolution than PAN image, thus were more applicable for color-based target discrimination task. The good fusion of the MS and PAN images is able to utilize the advantages of both which preserving the spectral resolution of MS images and spatial resolution of PAN image [16].

The UNB pan-sharp proposed in [17] is a fully automated one-step algorithm has been used in several commercial software e.g. PCI Geomatics, Digital Globe. The algorithm utilizes the least squares technique to find the best fit between the grey values of the PAN band and the MS bands to adjust the contribution of individual bands to the fusion result [14]. The influence of dataset variation in the fusion can be eliminated by a set of statistic approaches to establish the relationship between gray values of different bands. Therefore, without any user-specified parameters, UNB produce consistent high-quality fusion results regardless of sensor and image variation.

*B. Reflectance Contrast Analysis*

This section provides reflectance contrast index to analyze the ship discrimination performance of pan-sharpened MS bands in compare to PAN.

Reflectance contrast, or Target-Cluster-Ratio in radar image, is a widely used index to analyze the targets discrimination from the surrounding sea in both optical and Synthetic Aperture Radar [15], [18], [19]. Mean Reflectance Contrast (MRC) is defined as the ratio of the mean reflectance value of a ship target ($M_t$) to the mean reflectance of its surrounding region ($M_b$) while avoiding selection of other ships or land:

$$C_M = M_t/M_b \quad (1)$$

Besides, we introduce the First-Quartiles Reflectance Contrast (FQRC) index as follow:

$$C_{Q1} = Q1_t/M_b \quad (2)$$

where $Q1_t$ is the first quartile value of a ship target. While the magnitude of $C_M$ and $C_{Q1}$ both represent how well a band contrasts the ships from surrounding region, $C_{Q1}$ is introduced since the grey level of ship's pixels is vary. Fig. 2 shows the box-and-whisker plot of spectral values of two ships: one has homogeneous texture and the other has heterogeneous texture. As shown, the spectral values of the homogeneous ship cluster around its mean while the spectral values of heterogeneous ship are very spread out from the mean. Therefore, $C_{Q1}$ in combination with $C_M$ would be sufficient to assess how well the lowest contrast part of ship discriminates from surrounding region.

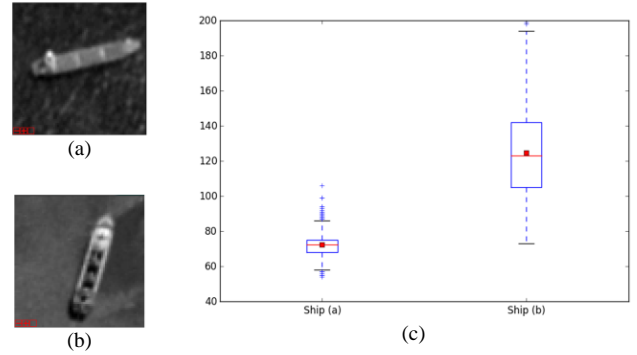

Fig. 2. Spectral comparison of PAN band between ships has homogeneous texture and heterogeneous texture. (a) Ship has homogeneous texture. (b) Ship has heterogeneous texture. (c) Box plots the pixel reflectance values of each ship.

Reflectance contrast may receive any positive value. A reflectance contrast of one indicates lack of contrast between pixels with and without ships. For the scene where there is appearance of interferences (clouds, ship wakes, etc.), the value can be less than one. It means that ships appear to be less bright than other interferences, thus extraction of ships may produce many false alarms while low contrast ships may be missing.

*C. Principle Component Analysis*

Using reflectance contrast index, bands which better discriminate ship from surrounding water are selected. To take the advantage of the different spectral bands, a data fusion method can be applied to combine the complementary information. Regarding data fusion, Principle Component Analysis (PCA) is utilized for creating a fused image which was found to achieve increased performance of ship target detection compared to single band detection performance. This section provides a brief description of PCA.

PCA first transform each multi-band image $A_{M \times N \times K}$, with $M$, $N$, $K$ denoting the number of rows, columns and bands respectively, into $M \times N$ vector of

$$x_i = [x_1, x_2, \dots, x_K]^T \quad (3)$$

with all pixel values $x_1, x_2, \dots, x_K$ at one corresponding pixel location of multi-band image data. The mean vector is calculated as:

$$m = \frac{1}{N \times M} \sum_{i=1}^{N \times M} [x_1, x_2, \dots, x_K]^T_i \quad (4)$$

The sample covariance matrix of $x$ is defined as

$$C_x = \frac{1}{N \times M} \sum_{i=1}^{N \times M} (x_i - m)(x_i - m)^T \quad (5)$$

We then find the eigenvectors $e_i$ with corresponding eigenvalues $\lambda_i$ of the sample covariance matrix, which satisfy:

$$C_x = ADA^T \tag{6}$$

where $D = diag(\lambda_1, \lambda_2, ..., \lambda_K)$ is the diagonal matrix composed of the eigenvalues $\lambda_1, \lambda_2, ..., \lambda_K$, and $A = (e_1, e_2, ..., e_K)$ is the orthonormal matrix composed of the corresponding $K$ dimension eigenvectors.

In [20], the authors indicated that the information content of PCA bands decreases with an increasing number of PCA bands, and most of the information may only be contained in the first few PCA bands. Since the proposed ship proposals extraction algorithm work on a single band input, the first PCA band which concentrates most of available information is used. Let the eigenvalues and eigenvectors be arranged in descending order so that $\lambda_1 > \lambda_2 > \cdots > \lambda_K$, thus the first row of the matrix $A^T$, namely the first eigenvector can be used to calculate the first PCA band as follow:

$$z_i = [z_1] = [e_{11} \quad e_{12} \quad \cdots \quad e_{1K}] \begin{bmatrix} x_1 \\ x_2 \\ \vdots \\ x_K \end{bmatrix} \tag{7}$$

where $z_i$ is the pixel value of the first PCA band at location $i$, $\begin{bmatrix} x_1 \\ x_2 \\ \vdots \\ x_K \end{bmatrix}$ is the pixel vector of original multi-band image at location $i$.

*D. Ship Proposals Extraction*

In optical image, ship can be viewed as abnormalities in open homogeneous ocean [2]. Therefore they can be detected by finding pixels of unusual brightness by comparing the encountered intensity with the statistical properties of the local and global sea background [19]. In this paper, we follow the approach proposed by [2] using global intensity anomaly and the regional texture anomaly. The texture anomaly suppression term is introduced to wipe out pixels appeared as white noise. From this perspective, the ships will be projected, while the clutters areas of sea will be suppressed. Finally, ship candidates can be extracted by finding a threshold on the linear combination of global intensity anomaly and local texture anomaly map.

*1) Global intensity anomaly*

Since the ship size is small and the major region of the image is open sea water, the intensity frequencies of pixels belong to ship target is usually very low [2]. Thus, the global anomaly is defined to emphasize the intensity anomaly of the ship as follow:

$$S_{global} = \frac{1}{f(x, y)} \tag{8}$$

where f(x, y) is the intensity frequency of the pixel (x, y).

*2) Regional texture anomaly*

Unlike global anomaly, the regional anomaly is a measure in a local region around the target. Ships can be regarded as singular structures in a noisy background, therefore generating conspicuous response to an edge detection method [21].

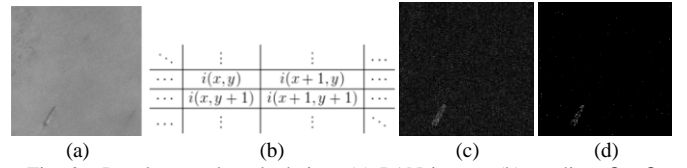

Fig. 3. Local anomaly calculation: (a) PAN image; (b) gradient $2 \times 2$ pixels mask; (c) gradient map; (d) gradient map after suppression

Given the image $i$, the intensity value at pixel $(x, y)$ is denoted as $i(x, y)$ (Fig. 3b). The image gradient magnitude is computed as:

$$g_x(x, y) = \frac{i(x + 1, y) + i(x + 1, y + 1) - i(x, y) - i(x, y + 1)}{2} \tag{9}$$

$$g_y(x, y) = \frac{i(x, y + 1) + i(x + 1, y + 1) - i(x, y) - i(x + 1, y)}{2} \tag{10}$$

and the regional anomaly as:

$$S_{regional} = G(x, y) = \sqrt{g_x^2(x, y) + g_y^2(x, y)} \tag{11}$$

Pixels with small gradient magnitude correspond to sea surface noise appear in dependence of the wave characteristics of the ocean. Also, they naturally present a higher error in the gradient computation due to the quantization of their values [22]–[24]. Thus, pixels with gradient magnitude $G(x, y)$ smaller than threshold $\rho$ should be discarded.

Gradient threshold $\rho$ is calculated follow the theorem in [25] which showed that gray-level quantification produces errors in the gradient orientation angle. When the gradient magnitude is large, this error is negligible, but it can be significant for s small gradient magnitude. The value of $\rho$ is set so that points where gradient magnitude larger than $\rho$, its angle error would be smaller than a pre-defined angle tolerance $\tau$.

Let $i, \tilde{\imath}$ and $n$ denote the image, the quantized image and the quantization noise respectively. We have:

$$\tilde{\imath} = i + n \tag{12}$$

$$\nabla \tilde{\imath} = \nabla i + \nabla n \tag{13}$$

In [25], the error of the image gradient angle can be bound by (see Fig. 4):

$$|angle\ error| = arcsin\left(\frac{q}{|\nabla i|}\right) \tag{14}$$

where $q$ is a bound to $|\nabla n|$. Imposing that $|angle\ error| \leq \tau$, we obtain:

$$\rho = \frac{q}{\sin \tau} \tag{15}$$

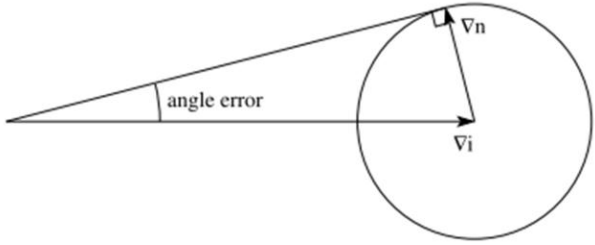

Fig. 4. Estimation of angle error due to quantization noise (adapted from [23]).

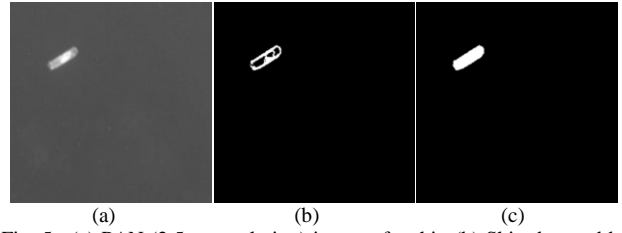

Fig. 5. (a) PAN (2.5 m resolution) image of a ship (b) Ship detected by Otsu's threshold (c) Ship target obtained after morphological opening operator.

The gradient magnitude threshold $\rho$ is set using (15) where $q$ is a bound on the possible error in the gradient value due to quantization effect and $\tau$ is the angle tolerance. For empirical reason, we set $q = 2$ and $\tau = 22.5$ degree as the one that gives the best results. Though these parameters have the same values as proposed in [23], [26], there is no theory behind. Fig. 3c and Fig. 3d show the results of texture suppression where small magnitude edges generated by noisy background water are removed, while large magnitude edges emphasize edges of the ship are remain after suppression.

*3) Linear combination*

The global and regional contrasts are integrated as follow:

$$S_{final} = S_{regional} + S_{global} \quad (16)$$

where $S_{regional}$ and $S_{global}$ are both normalized into range [0,1]. By using integrated global-regional anomaly map, we guarantee that the target proposal must meet two conditions as the same time. It can effectively suppress both large region (unusual water pattern, cloud cover, etc.) and noise generated by heterogeneous texture of sea surface.

*4) Threshold-based segmentation*

In image processing, gray-level thresholding is one of the most commonly used approach for foreground and background segmentation due to its simple-ness and effectiveness [27], [28]. In the past decades, various thresholding methods, categorized by the information they are exploiting, have been developed [29]. However, they are usually unable to generate satisfactory segmentation results in the case that images are corrupted by noise as all noise and edges are labeled as foreground or background [28]. In this article, five methods from four categories are selected to perform thresholding on calculated abnormality map. As the scope of our research is not to identify the best threshold algorithm, we intent to use the segmentation results to demonstrate how texture suppression help increase the robustness in regards of thresholding methods. Brief descriptions of these thresholding methods are introduced as follow.

Global clustering-based: Otsu's [30] and IsoData [31], [32] are used. Otsu's method assumes that the image contains two classes of pixels following bi-modal histogram (foreground pixels and background pixels). It then calculates the optimum threshold separating the two classes so that their combined spread (intra-class variance) is minimal, or equivalently (because the sum of pairwise squared distances is constant) so that their inter-class variance is maximal [30]. In similar study, IsoData first divides the image into foreground and background region by taking an initial threshold. Then, an iterative procedure, which increases the threshold according to the computed averages of foreground and background pixels, is repeated until the threshold is larger than the composite average.

Global entropy-based: Yen's method [27] find the threshold that maximize the entropy of the distribution of gray levels. It can be interpret as indicative of maximum information transfer [29].

Global histogram-based: mean method [33] uses the mean of grey levels as the threshold.

Local adaptive: a threshold is calculated at each pixel, which depends on some local statistics. Sauvona's method [34] adapts the threshold according to the local mean and standard deviation of $b \times b$ pixels size window around each pixel.

*5) Morphological operator*

Finally, morphological opening operator, defined as erosion followed by dilation, with a $3 \times 3$ pixels structural element, and *iteration* parameter set to 2 is applied on threshold image. Since ship target is a cluster of bright pixels, isolated pixels resulting from image thresholding are considered as noise and are removed in erosion stage. As shown in Fig. 5, not every part of the ship can be detected during thresholding operation due to the variation of intensity inside ship target. The dilation stage act as region growing operator allows the grouping of the separated regions.

*E. False Alarm Elimination*

Based on a priori knowledge of ship's shape characteristics, we described ship target by three simple, low computing complexity features: *width*, *length* and *length-width-ratio*. *width* and *length* are measured as the short side and long side of the minimum rotated bounding box of the region respectively. *length-width-ratio* is calculated as follow:

$$length - width - ratio = length/width \quad (17)$$

After false alarm elimination, the resulting regions are returned as detected ship proposals.

*F. Metric Evaluation and Validation*

To quantify the evaluation, we label ship bodies manually in all images as the ground truth. A detected proposal (DP) is marked as true positive if there is a ground truth (GT) satisfied:

$$IoU = \frac{Area(GT \cap DP)}{Area(GT \cup DP)} > T \quad (18)$$

where $T$ is a threshold value in range $[0, 1]$. Intersection over Union (IoU) is used to compute the intersection of the detected proposal and the ground truth target divided by the area of their union [35].

For a given IoU threshold $T = 0.5$, the numbers of the detected ships, missed ships, and false alarms ships are counted to compute the recall rate, precision and F1-score as follow:

$$recall\ rate_{T=0.5} = \frac{detected\ ships}{detected\ ships + missed\ ships} \quad (19)$$

$$precision_{T=0.5} = \frac{detected\ ships}{detected\ ships + false\ alarm\ ships} \quad (20)$$

$$f1\_score_{T=0.5} = 2 \times \frac{recall\ rate_{T=0.5} \times precision_{T=0.5}}{recall\ rate_{T=0.5} + precision_{T=0.5}} \quad (21)$$

Beside, we also report the average recall (AR) [36] by averaging the recall rate for $T \in [0.5, 1]$. Due to its ability to summarizes proposal performance across thresholds $T$, AR is proved to be highly correlated with the performance of detector used in the later stage.

## III. EXPERIMENTAL RESULTS AND DISCUSSION

### A. Study Area and Datasets

In this paper, VNREDSat-1 optical image is used to evaluate the proposed approach. VNREDSat-1, which was successfully launched on May 7, 2013, is the first high-resolution Earth observation system satellite of Vietnam. It carries two panchromatic/multi-spectral cameras with 2.5-meter and 10-meter resolutions respectively.

11 VNREDSat-1 satellite scenes taken under different illuminations and contain different clutters (clouds, waves, etc.) are collected covering two study areas including Saigon River and South China Sea. All MS images are pan-sharpened to 2.5-meter spatial-resolution follow the UNB method. 11 images are subdivided into 105 sub-images with $256 \times 256$ pixels size contain 166 ship targets of different sizes and shapes. All ships are on-screen labeled using PAN and a color composite of near-infrared (red), green (blue), red (green). Two Dataset are derived as follows:

*1) Dataset1*

105 sub-images are divided into two groups which are clear water (CW) and turbid water (TW). With regards to turbidity, the results from [37]–[39] showed that the water in Saigon River had high turbidity, which exceeded the Vietnamese technical regulation for surface water (5 Nephelometric Turbidity Units), because of impacts of navigation and urbanization. Therefore, 33 sub-images containing 58 ships covering Saigon River area are classified in TW group. Meanwhile, the rest of 72 sub-images containing 108 ships taken in deep ocean area of South China Sea are considered in CW group. The *Dataset1* is used to analyze the ship discrimination performance of pan-sharpened MS in compare to PAN in different turbidity level of water.

*2) Dataset2*

To evaluate the performances of the proposed ship proposal detection approach in different cases, all ground truth ships are grouped into *Low Contrast* and *High Contrast* based on their MCR value retrieved from PAN band. The number of ship target in each group, detailed by a range of 0.25 MCR value, is shown in TABLE I. Only groups which have sample ship are listed.

TABLE I
NUMBER OF SHIP TARGETS IN EACH GROUP OF *DATASET2*.

| | Low contrast | | High contrast | | | | | | |
|---|---|---|---|---|---|---|---|---|---|
| MCR range | 0.75 - 1 | 1 - 1.25 | 1.25 - 1.5 | 1.5 - 1.75 | 1.75 - 2 | 2 - 2.25 | 2.25 - 2.5 | 2.5 - 2.75 | 3 - 3.25 |
| #num of ship | 35 | 52 | 25 | 22 | 17 | 7 | 4 | 3 | 1 |

The first and the second row indicate the range of reflectance contrast of each group retrieved from PAN band. The last row shows the number of ground truth ships.

### B. Reflectance Contrast Analysis

Fig. 6 shows the histogram of the Mean Reflectance Contrast and First-Quartile Reflectance Contrast for PAN and four MS bands. TABLE II shows the mean reflectance contrast of each band in clear water in turbid water.

These results revealed a significant difference in mean reflectance contrast between five bands in clear water. The near infrared band of VNREDSat-1 VHR optical image best discriminated ships ($Mean\ MRC = 1.969$; $Mean\ FQRC = 1.638$), followed by Red ($Mean\ MRC = 1.698$; $Mean\ FQRC = 1.426$) and PAN ($Mean\ MRC = 1.551$; $Mean\ FQRC = 1.306$) bands, while the Blue and Green band poorly discriminated the ships. It can be explained by that pure water absorbs almost all of the incident near infrared radiant flux [40] while ship object reflects significant amount of near infrared energy, thus water appears darker on this band. These results agree with those of [41] who concluded that Landsat TM red (band 3) and near infrared (band 4) bands were the most useful for ship detection in marine clear water environments.

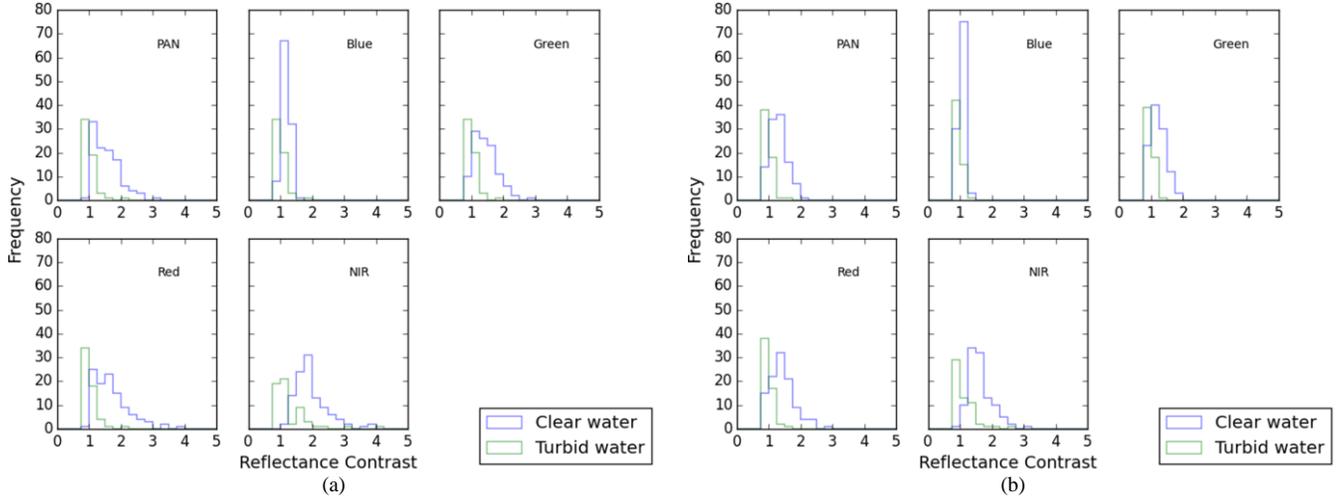
Fig. 6. Histogram of the (a) Mean Reflectance Contrast (MRC) and (b) First-Quartile Reflectance Contrast (FQRC) of ships for PAN and pan-sharpened MS bands

TABLE II
REFLECTANCE CONTRAST OF SHIPS AND SURROUNDING REGION FOR VNREDSAT-1 BANDS

| Bands | Clear water | | Turbid water | |
|---|---|---|---|---|
| | Mean MRC | Mean FQRC | Mean MRC | Mean FQRC |
| PAN | 1.551 | 1.306 | 1.066 | 1.013 |
| Blue | 1.179 | 1.052 | 1.038 | 0.980 |
| Green | 1.434 | 1.224 | 1.043 | 0.985 |
| Red | 1.698 | 1.426 | 1.080 | 1.016 |
| NIR | 1.969 | 1.638 | 1.286 | 1.150 |

The ship discrimination performances of all bands reduce dramatically on turbid water. TABLE II shows that the reflectance contrasts of all bands centres almost on one. It can be explained by the presence of organic and inorganic constituents in the near surface of turbid water column. These materials cause the peak reflectance shifts toward longer wavelength in the visible and near infrared region [40]. Thus, turbid water has high reflectance in the visible and near infrared bands which results in a lower reflectance contrast of ships. Our results is consistent with the result of [15], [42] which shows that the visible and near infrared bands of Landsat TM better discriminate ship from clear water than turbid water. As can be seen in TABLE II, only NIR band has noticeable reflectance contrast of 1.286 for Mean MRC and 1.150 for Mean FQRC. Though the contrast is not as significant as in clear water, we will show that NIR can boost up the detection rate of ship in turbid water in the next section.

*C. Ship Proposals Detection*

In this part, we validate the performance of our ship proposal detection method on *Dataset2*. First, we report the overall detection performance of proposed method using recall rate and average recall. Second, we compare our method with other state-of-the-art anomaly-based models to demonstrate the effectiveness and robustness of our proposed method. Finally, we compare the abnormality map generated by our proposed model with those of other models to analyze why our proposed models has better results.

*1) Detection performance*

As shown in previous section, NIR band has been proved to be superior for ship discrimination. While PAN, Red and Green band both show significant performance in clear water, only PAN band is selected. This can be explained by that the wavelength of PAN spanning a large part of the visible part of the spectrum. PCA component of NIR and PAN image are used as the input of our proposed model. Finally, several automatic threshold selection methods described in Section II will be applied to separate foreground ship from background.

Fig. 7a shows the results of our proposed anomaly model with different thresholding techniques on high contrast dataset. Our method achieved highest $recall\ rate_{T=0.5}$ of 91.14% and average recall of 45.9% using Yen's threshold. It's not surprise that Otsu's and Isodata method provide similar performance (86.07% and 87.34% of $recall\ rate_{T=0.5}$ and 43.9% and 44.1% of average recall respectively) since both are clustering based. Mean method provides worse results in term of both $recall\ rate_{T=0.5}$ (81.01%) and average recall (39.6%) since it simply uses mean value as threshold.

Meanwhile, Fig. 7b shows that the results of our proposed anomaly model decrease dramatically on low contrast dataset. Yen's, Otsu's and Isodata share the same $recall\ rate_{T=0.5}$ of 43.68% with little to no different on average recall value (45.9%, 43.9% and 44.1% respectively). Mean method still provides worse results (41.38% of $recall\ rate_{T=0.5}$ and 39.6% of average recall).

The performance results on both high and low contrast dataset show that, despite there is some difference, the detection performance resulted by different thresholding techniques are comparable and consistent across dataset. Interestingly, simple threshold techniques such as Mean's and Sauvona's can work well with anomaly map generated by our proposed model. It proves that our proposed method is not affected by how we choose thresholding technique.

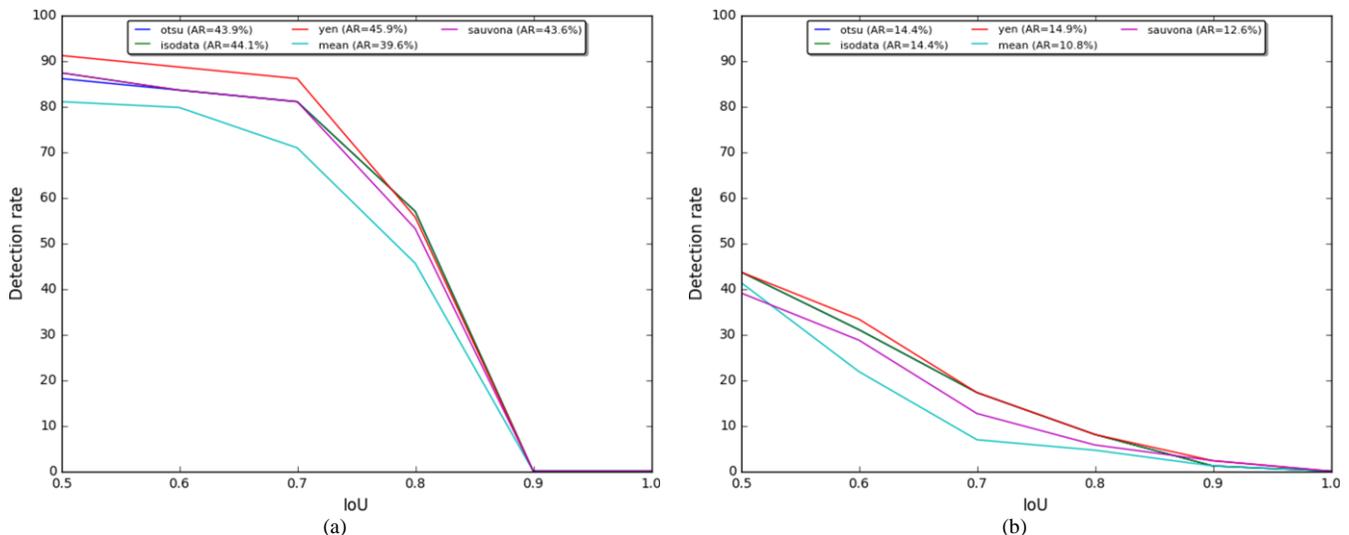

Fig. 7. Comparative experiment results of our proposed model using different threshold techniques on (a) High Contrast Dataset and (b) Low Contrast Dataset.

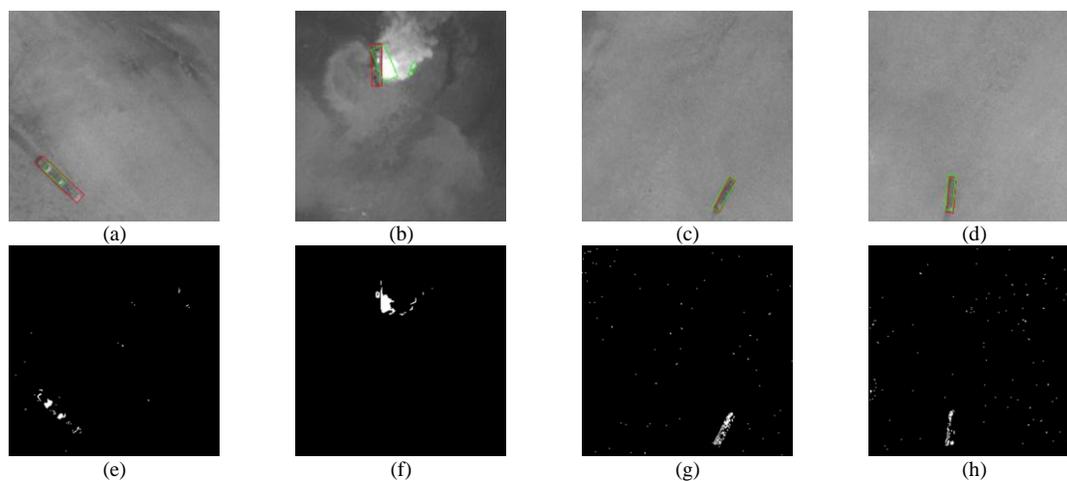

Fig. 8. Results of ship on *Dataset2-Low Contrast* dataset. The first row is the PAN image with ground truth (red rectangle) and detected segment (green rectangle) rotated bounding box. The second row shows their segmentation image respectively. The ships in (a) and (b) cannot be detected. The ships in (c) and (d) are detected correctly.

Fig. 8 shows some results of our method on *Dataset2-Low Contrast* dataset. As can be seen, some ships can be detected correctly but the missing rate is high. There are two reasons. The first one is that the ship or part of it has very low contrast to surround water that they can hardly be identified even for human. Another reason is the appearance of interferences adjoining ships such as ship wake, clouds that has very high contrast. Thus, traditional segmentation methods used in this article fail to distinguish them from ship.

*2) Comparisons with other approaches*

We compare our method with the state-of-the-art anomaly-based ship detection models including Corbane et al. 2008 [4], Yang et al. 2014 [2], Shi et al. 2014 [5] and the method that utilize widely used Reed-Xiaoli Detector (RXD) algorithm [6]. Note that we only compare the part of ship proposals extraction of those methods. Corbane's [4] and Shi's [5] method both provided a strong classifier for false alarm elimination which is out of scope of this paper. Additionally, to see how NIR data and texture anomaly suppression term contribute to the final performance, we also compare our proposed model (PANIR-WTAS) with three other variant: 1) our model with texture anomaly suppression term on PAN (PAN-WTAS); 2) our model without texture anomaly suppression term on PCA of NIR and PAN (PANIR-WOTAS); and 3) our model without texture anomaly suppression term on PAN (PAN-WOTAS). All experiments are setup as follow. Corbane et al. 2008 [4], Yang et al. 2014 [2], Shi et al. 2014 [5] require only original PAN image as input. Since those methods require several manual configurations which vary on different dataset, we follow suggested parameters on their papers. For the method that utilize widely used Reed-Xiaoli Detector (RXD) algorithm [6], we calculate anomaly map using PAN and NIR data.

In this section, the results of all models using Mean's and Sauvona's threshold are excluded, regarding the fact that these thresholding methods are too simple and not widely used in researches. Detail performance of models on all threshold methods including Mean's and Sauvona's can be seen in

TABLE III
AVERAGE RECALL OVERALL DATASET2 USING DIFFERENT THRESHOLD METHODS

|  | Corbane et al. | Yang et al. | Shi et al. | RXD | PAN-WOTAS | PANIR-WOTAS | PAN-WTAS | PANIR-WTAS |
|---|---|---|---|---|---|---|---|---|
| Otsu's | 20.08 | 21.89 | 3.21 | 11.04 | 32.13 | 31.63 | 31.53 | 33.13 |
| IsoData | 14.36 | 21.59 | 14.26 | 11.14 | 13.56 | 22.69 | 31.33 | 33.23 |
| Yen's | 30.02 | 2.71 | 8.43 | 20.38 | 32.63 | 18.98 | 32.43 | 34.54 |
| #mean+SD | 21.49 (±7.92) | 15.40 (±10.99) | 8.63 (±5.53) | 14.19 (±5.36) | 26.11 (±10.87) | 24.43 (±6.50) | 31.76 (±0.59) | 33.63 (±0.79) |

The last row shows the mean and standard deviation of average recall (*#mean+SD*) of each anomaly-based model across different threshold method.

TABLE IV
AVERAGE RECALL ON THE DATASET2-HIGH CONTRAST USING DIFFERENT THRESHOLD METHODS

|  | Corbane et al. | Yang et al. | Shi et al. | RXD | PAN-WOTAS | PANIR-WOTAS | PAN-WTAS | PANIR-WTAS |
|---|---|---|---|---|---|---|---|---|
| Otsu's | 31.83 | 31.65 | 5.79 | 13.56 | 46.11 | 43.94 | 46.47 | 43.76 |
| IsoData's | 24.23 | 31.10 | 23.69 | 13.38 | 19.53 | 31.28 | 46.47 | 43.94 |
| Yen's | 43.94 | 4.16 | 10.13 | 27.67 | 42.13 | 31.83 | 48.10 | 45.75 |
| #mean+SD | 33.33 (±9.94) | 22.30 (±15.71) | 13.20 (±9.34) | 18.20 (±8.20) | 35.92 (±14.34) | 35.68 (±7.16) | 47.01 (±0.94) | 44.48 (±1.10) |

The last row shows the mean and standard deviation of average recall (*#mean+SD*) of each anomaly-based model across different threshold method.

TABLE V
AVERAGE RECALL ON THE DATASET2-LOW CONTRAST USING DIFFERENT THRESHOLD METHODS.

|  | Corbane et al. | Yang et al. | Shi et al. | RXD | PAN-WOTAS | PANIR-WOTAS | PAN-WTAS | PANIR-WTAS |
|---|---|---|---|---|---|---|---|---|
| Otsu's | 3.94 | 7.06 | 0.00 | 5.75 | 10.67 | 11.82 | 9.36 | 14.45 |
| IsoData | 1.48 | 7.06 | 1.81 | 6.08 | 4.43 | 8.70 | 9.03 | 14.45 |
| Yen's | 9.20 | 0.66 | 4.60 | 8.21 | 15.11 | 2.13 | 9.36 | 14.94 |
| #mean+SD | 4.87 (±3.94) | 4.93 (±3.70) | 2.14 (±2.32) | 6.68 (±1.34) | 10.07 (±5.37) | 7.55 (±4.95) | 9.25 (±0.19) | 14.61 (±0.28) |

The last row shows the mean and standard deviation of average recall (*#mean+SD*) of each anomaly-based model across different threshold method.

Appendix.

TABLE IV and TABLE V show the detection performance of compared methods on *Dataset2* high and low contrast respectively. In high contrast dataset, PAN-WTAS is slightly better than PANIR-WTAS with mean average recall of 47.01% and 44.08% respectively. Meanwhile, in the low contrast dataset, PANIR-WTAS outperforms PAN-WTAS (14.61% and 9.25% mean average recall). Several interesting find out can be drawn as follow.

Firstly, in both high contrast and low contrast dataset, the average recall of Corbane et al. 2008 [4], Yang et al. 2014 [2], Shi et al. 2014 [5], RXD [6], PAN-WOTAS and PANIR-WOTAS is inconsistent across different threshold techniques. There is a huge gap between the performance of these models on best-case and worst-case threshold (for example, Corbane et al. 2008 [4] achieved highest average recall of 43.94% using Yen's threshold, while decrease dramatically to 24.23% using IsoData's threshold on *Dataset2-High Contrast*). We address this phenomenon using *#mean+SD* values showed in the last row of TABLE IV and TABLE V. As can be seen, the average recall of Corbane et al. 2008 [4], Yang et al. 2014 [2], Shi et al. 2014 [5], RXD [6], PAN-WOTAS and PANIR-WOTAS have much higher standard deviation (*SD*) than PAN-WTAS and PANIR-WTAS. Thus, for these models to be able to work, the trial-error process is needed to find out the best working threshold.

Secondly, though ships have better reflectance contrast in NIR than PAN, the usage of NIR does not always improve the recall rate of ship detection task. In high contrast dataset, both two models PANIR-WOTAS and PANIR-WTAS that incorporated NIR data have lower mean average recall than their respective model that only use PAN data (PAN-WOTAS and PAN-WTAS). This can be explained by that, in high contrast scene, PAN is enough to discriminate ships from background sea surface (recall from Section III.B that, in clear water PAN have $Mean\,MRC$ of 1.551). While introducing spectral noise as the result of pan-sharpen process, the incorporation of NIR may not help increase the separation of ship from background. The similar issue can be found in low contrast dataset as PANIR-WOTAS has lower mean average recall than PAN-WOTAS. However, in this low contrast dataset, the usage of NIR together with texture anomaly suppression can help improve the recall rate of the model. It should be note that, PANIR-WTAS not only achieves highest mean average recall of 14.61% but also has little to no different in performance using Otsu', IsoData's or Yen's

TABLE VI
THE $f1\_score_{T=0.5}$ OF EACH ANOMALY-BASED MODEL USING DIFFERENT THRESHOLDS.

|  | Corbane et al. | Yang et al. | Shi et al. | RXD | PAN-WOTAS | PANIR-WOTAS | PAN-WTAS | PANIR-WTAS |
|---|---|---|---|---|---|---|---|---|
| Otsu's | 38.59 | 52.77 | 12.7 | 34.71 | 58.72 | 62.91 | 60.31 | 63.48 |
| IsoData | 22.38 | 45.06 | 44.44 | 34.18 | 30.66 | 45.37 | 60.62 | 64.07 |
| Yen's | 49.6 | 4.68 | 26.96 | 45.84 | 62.3 | 23.53 | 59.06 | 62.15 |
| #mean+SD | 36.86 ($\pm$13.69) | 34.17 ($\pm$25.83) | 28.03 ($\pm$15.9) | 38.24 ($\pm$6.58) | 50.56 ($\pm$17.33) | 43.94 ($\pm$19.73) | 60 ($\pm$0.83) | 63.23 ($\pm$0.98) |

The last row shows the mean and standard deviation of average recall (*#mean+SD*) of each anomaly-based model across different threshold method.

threshold.

TABLE III shows the detection performance overall *Dataset2*. As can be seen, our proposed model (PANIR-WTAS) not only have highest mean recall rate but also have lowest performance variation between different threshold. Besides, precision is another important factor to be considered. Ideally, we would like our model to mark all true ships that exist while exclude other interferences from candidate list. Then we would have both high recall rate and high precision. In TABLE VI, we report the F1 score calculated at $IoU = 0.5$ ($f1\_score_{T=0.5}$) of each anomaly-based model on *Dataset2*. Note that in this paper, we do not employ a strong classifier but a simple linear false alarm eliminator using only ship's *width*, *length* and *length-width-ratio* to classify proposal as ship or not.

Overall, our proposed model PANIR-WTAS, which have mean $f1\_score_{T=0.5}$, outperforms other models by at least 3%. Without texture anomaly suppression, NIR data may boost up the performance if we could find of suitable automatic threshold (PANIR-WOTAS with Otsu's) but may fail to other threshold (PANIR-WOTAS with Yen's). This pattern is also true for other models. Corbane et al. 2008 [4] works well with Yen's but doesn't work well with IsoData. The performance of Yang et al. 2014 [2] decreases dramatically from Otsu's ($f1\_score_{T=0.5} = 52.77$) to Yen's ($f1\_score_{T=0.5} = 4.68$). Shi et al. 2014 [5] achieves highest $f1\_score_{T=0.5}$ of 44.44 while fail to work with Otsu's threshold ($f1\_score_{T=0.5} = 12.7$). Among those RXD [6] is more stable which has lower variation in performance across different thresholds (6.58 SD of $f1\_score_{T=0.5}$). In conclusion, PANIR-WTAS has better performance both in term of F1 score value and its variation.

*3) Comparisons of anomaly maps*

In this section, we analyze the reason behind why our proposed anomaly-based model is robustness to threshold methods. Fig. 9 and Fig. 10 shows the anomaly image and their histogram generated by anomaly detectors under different background conditions. The simple background image represents the case of quite sea with no interferences and the ship is highly contrast to the surrounding water. Meanwhile, in the complex background image, the ship appeared as low contrast to the background due to the existent of interferences (e.g. ship wakes, sea clutter, illumination noise). The histogram of anomaly image is separated into two tails accordingly to threshold determined by Otsu's method. It helps us to better understand how different anomaly map will affect the way thresholding technique choose the threshold value.

As seen in Fig. 9, the performance of different anomaly detectors is similar for the images with a simple background. Since there is little to no interferences, the histogram of generated anomaly maps all form a bimodal-like shape in which Otsu's method can select good threshold to separate the ship from background.

However, our proposed method PANIR-WTAS outperforms the others over the images with complex background. In Fig. 10, for PANIR-WOTAS, Corbane et al. 2008, Shi et al. 2014, Yang et al. 2014, RXD with PAN+NIR, we can see that the anomaly values of noisy pixels may spread a wide range from the minimum to near maximum of anomaly values. Their anomaly image forms a unimodal distribution and as consequent, Otsu's method cannot select satisfactory threshold value (in these case, Otsu's method tends to select lower threshold value). Therefore, many pixels generated by interferences may be wrongly assigned to foreground object. In contrast, anomaly image generated by our proposed method PANIR-WTAS get better separation results due to the ability of suppressing the background interferences. It helps avoids the noises, therefore avoid the labeling of noisy pixels as foreground.

## IV. CONCLUSIONS

In this article, we proposed a novel framework for detecting and extracting ship proposals from optical remote sensing images, which includes fusion of pan-sharpened MS data and improved anomaly detector. The principle results obtained can be summarized as follow. First, to highlight ship target from surrounding background, pan-sharpened near infrared data is incorporated. Second, texture suppression term is applied to anomaly detector to wide out pixels appeared as white noise. A comprehensive analysis was performed shows that our proposed model cannot only achieved high recall but also robust to noise and produces stable results across different threshold methods.

Experiment results on VHR VNREDSat-1 satellite images in two study areas including Saigon River (turbid water) and South China Sea (clear water) show that 1) Ship pixels have higher reflectance contrast in pan-sharpen NIR than PAN and other visible bands, especially in clear water. In turbid water, ships have much lower contrast which usually similar to sea surface water. 2) Though P-NIR data can help discriminate ships from other interferences better, it may produce spectral noise to the image due to pan sharpen process, then as a result decrease the ship detection performance. 3) Our proposed model incorporated anomaly suppression term can help reduce the negative effect of noise. The results show that, our model not only achieves highest performance but also robust to different threshold methods.

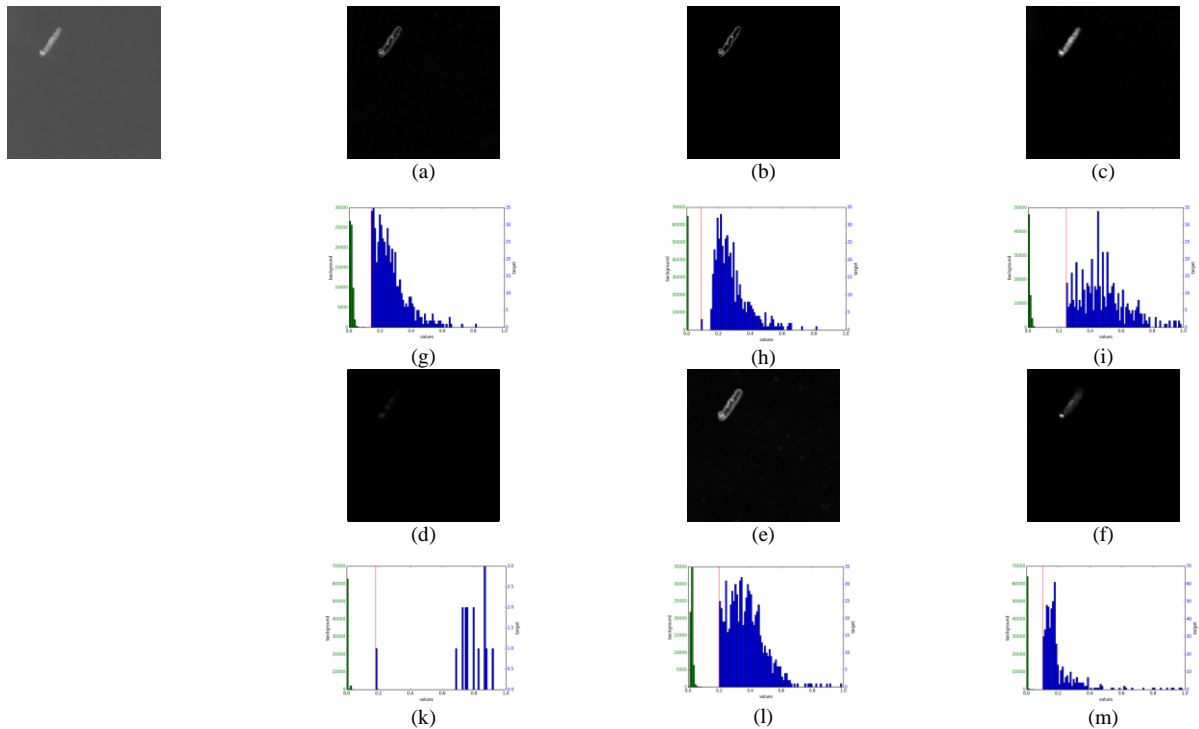

Fig. 9. Comparison of generated anomaly images in simple background scene: (a) PANIR-WOTAS; (b) PANIR -WTAS; (c) Corbane et al. 2008; (d) Shi et al. 2014; e) Yang et al. 2014; (f) RXD with PAN+NIR; and (g), (h), (i), (k), (l), (m) histogram of (a), (b), (c), (d), (e), (f) respectively. The red line indicates the threshold selected by Otsu's method.

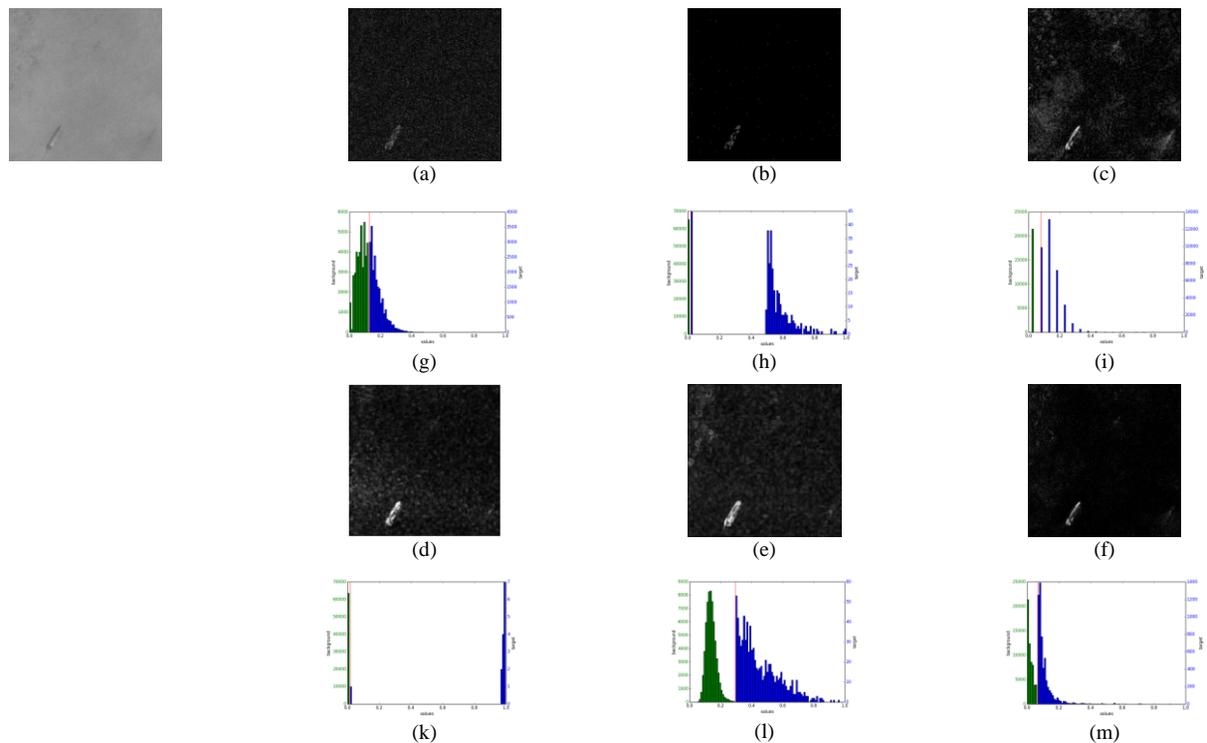

Fig. 10. Comparison of generated anomaly images complex background scene: (a) PANIR-WOTAS; (b) PANIR-WTAS; (c) Corbane et al. 2008; (d) Shi et al. 2014; e) Yang et al. 2014; (f) RXD with PAN+NIR; and (g), (h), (i), (k), (l), (m) histogram of (a), (b), (c), (d), (e), (f) respectively. The red line indicates the threshold selected by Otsu's method.

However, our result of the ship proposal detection performance is still not significant in case of ship with very low-contrast on turbid water. The dramatically decrease in accuracy suggest that traditional anomaly-based detector is limited on providing accurate pixel-level segment. Thus, our main objective will be to improve the segmentation performance using novel technique such as deep convolutional neural network.

APPENDIX

TABLE VII
AVERAGE RECALL ON THE *Dataset2-High Contrast* USING DIFFERENT THRESHOLD METHODS

|  | Corbane et al. | Yang et al. | Shi et al. | RXD | PAN-WOTAS | PANIR-WOTAS | PAN-WTAS | PANIR-WTAS |
|---|---|---|---|---|---|---|---|---|
| Otsu's | 31.83 | 31.65 | 5.79 | 13.56 | 46.11 | 43.94 | 46.47 | 43.76 |
| IsoData's | 24.23 | 31.10 | 23.69 | 13.38 | 19.53 | 31.28 | 46.47 | 43.94 |
| Yen's | 43.94 | 4.16 | 10.13 | 27.67 | 42.13 | 31.83 | 48.10 | 45.75 |
| Mean's | 2.53 | 0.36 | 4.7 | 0.36 | 0.00 | 0.00 | 43.76 | 39.60 |
| Sauvona's | 0.00 | 0.36 | 0.90 | 12.66 | 0.00 | 0.00 | 44.30 | 43.58 |
| #mean+SD | 20.51 (±18.94) | 13.53 (±16.37) | 9.04 (±8.82) | 13.53 (±9.67) | 21.55 (±22.13) | 21.41 (±20.19) | 45.82 (±1.77) | 43.33 (±2.26) |

TABLE VIII
AVERAGE RECALL ON THE *Dataset2-Low Contrast* USING DIFFERENT THRESHOLD METHODS

|  | Corbane et al. | Yang et al. | Shi et al. | RXD | PAN-WOTAS | PANIR-WOTAS | PAN-WTAS | PANIR-WTAS |
|---|---|---|---|---|---|---|---|---|
| Otsu's | 3.94 | 7.06 | 0.00 | 5.75 | 10.67 | 11.82 | 9.36 | 14.45 |
| IsoData | 1.48 | 7.06 | 1.81 | 6.08 | 4.43 | 8.70 | 9.03 | 14.45 |
| Yen's | 9.20 | 0.66 | 4.60 | 8.21 | 15.11 | 2.13 | 9.36 | 14.94 |
| Mean | 0.00 | 0.33 | 2.13 | 0.33 | 0.00 | 0.00 | 11.99 | 10.84 |
| Sauvona | 0.00 | 0.00 | 0.00 | 0.82 | 0.00 | 0.00 | 9.36 | 12.64 |
| #mean+SD | 2.92 (+-3.86) | 3.02 (+-3.69) | 1.71 (+-1.9) | 4.24 (+-3.48) | 6.04 (+-6.69) | 4.53 (+-5.42) | 9.82 (+-1.22) | 13.46 (+-1.71) |

TABLE IX
AVERAGE RECALL OVERALL *Dataset2* USING DIFFERENT THRESHOLD METHODS

|  | Corbane et al. | Yang et al. | Shi et al. | RXD | PAN-WOTAS | PANIR-WOTAS | PAN-WTAS | PANIR-WTAS |
|---|---|---|---|---|---|---|---|---|
| Otsu's | 20.08 | 21.89 | 3.21 | 11.04 | 32.13 | 31.63 | 31.53 | 33.13 |
| IsoData | 14.36 | 21.59 | 14.26 | 11.14 | 13.56 | 22.69 | 31.33 | 33.23 |
| Yen's | 30.02 | 2.71 | 8.43 | 20.38 | 32.63 | 18.98 | 32.43 | 34.54 |
| Mean's | 1.4 | 0.4 | 3.92 | 0.4 | 0 | 0 | 31.63 | 28.62 |
| Sauvona's | 0 | 0.2 | 0.5 | 7.53 | 0 | 0 | 30.32 | 31.93 |
| #mean+SD | 13.17 (±12.7) | 9.36 (±11.35) | 6.06 (±5.4) | 10.1 (±7.22) | 15.66 (±16.23) | 14.66 (±14.15) | 31.45 (±0.76) | 32.29 (±2.25) |

TABLE X
THE $f1\_score_{T=0.5}$ OF EACH ANOMALY-BASED MODEL USING DIFFERENT THRESHOLD

|  | Corbane et al. | Yang et al. | Shi et al. | RXD | PAN-WOTAS | PANIR-WOTAS | PAN-WTAS | PANIR-WTAS |
|---|---|---|---|---|---|---|---|---|
| Otsu's | 38.59 | 52.77 | 12.7 | 34.71 | 58.72 | 62.91 | 60.31 | 63.48 |
| IsoData | 22.38 | 45.06 | 44.44 | 34.18 | 30.66 | 45.37 | 60.62 | 64.07 |
| Yen's | 49.6 | 4.68 | 26.96 | 45.84 | 62.3 | 23.53 | 59.06 | 62.15 |
| Mean | 2.78 | 0.73 | 10.44 | 1.15 | NaN | NaN | 50.59 | 48.9 |
| Sauvona | NaN | 0.69 | 0.73 | 19.51 | NaN | NaN | 39.75 | 42.56 |